\documentclass[lettersize,journal]{IEEEtran}

\usepackage{amsmath,amsfonts}
\usepackage{algorithmic}
\usepackage{algorithm}
\usepackage{array}
\usepackage[caption=false,font=normalsize,labelfont=sf,textfont=sf]{subfig}
\usepackage{textcomp}
\usepackage{stfloats}
\usepackage{url}
\usepackage{tabularx}
\usepackage{verbatim}
\usepackage{booktabs}
\usepackage{graphicx}
\usepackage{xcolor}
\usepackage{geometry}

\usepackage{cite}
\usepackage{enumitem} 
\begin{document}

\title{Strategic Application of AIGC for UAV Trajectory Design: A Channel Knowledge Map Approach}

\author{Chiya Zhang, Ting Wang, Rubing Han, Yuanxiang Gong
\thanks{The authors are with School of Electronic and Information Engineering, Harbin Institute of Technology, Shenzhen, China.} 
\thanks{This work was supported by the NSFC Fund 62394294, U20A20156 and 62101161. This work was also supported by Foundation of National Key Laboratory of Radar Signal Processing under Grant JKW202303.

\textit{Corresponding author: Ting Wang(24B352015@stu.hit.edu.cn)}}}


\maketitle

\begin{abstract}
Unmanned Aerial Vehicles (UAVs) are increasingly utilized in wireless communication, yet accurate channel loss prediction remains a significant challenge, limiting resource optimization performance. To address this issue, this paper leverages Artificial Intelligence Generated Content (AIGC) for the efficient construction of Channel Knowledge Maps (CKM) and UAV trajectory design. Given the time-consuming nature of channel data collection, AI techniques are employed in a Wasserstein Generative Adversarial Network (WGAN) to extract environmental features and augment the data. Experiment results demonstrate the effectiveness of the proposed framework in improving CKM construction accuracy. Moreover, integrating CKM into UAV trajectory planning reduces channel gain uncertainty, demonstrating its potential to enhance wireless communication efficiency. 
\end{abstract}
\def\abstractname{Note to Practitioners}
\begin{abstract}
This paper addresses the challenge of accurate channel loss prediction in UAV-enabled wireless communication. Traditional methods for channel prediction are often limited by the time-intensive nature of data collection and environmental variability. An AI-driven approach using a WGAN to enhance Channel Knowledge Maps (CKM) by efficiently augmenting channel data with high-quality synthetic data is introduced. Implementing this approach can reduce uncertainty in channel gain prediction, leading to more reliable and robust wireless communication. Developments in AIGC can further enhance these techniques, offering even greater accuracy and efficiency in channel prediction.
\end{abstract}

\section{Introduction}
\IEEEPARstart{U}{AVs} can supplement existing wireless networks to provide high-quality service and, under certain conditions, can serve as base stations. In scenarios where traditional communication infrastructure is compromised or overloaded, UAVs can ensure continuous connectivity and coverage. Equipped with sensors, cameras, and other devices, they can effectively establish line-of-sight channels in high-altitude scenarios and cover extensive areas. As UAV trajectories design has emerged as a key topic in wireless communications research, various work on this topic have been investigated\cite{ref1}. The channel model between UAVs and ground users is a crucial factor in ensuring reliable and efficient communication within UAV-IoT networks. Typically, this model can be seen as a combination of line-of-sight (LoS) and non-line-of-sight (NLoS) conditions\cite{ref2}. Advanced models can achieve greater accuracy by using three-dimensional (3D) building and terrain data, with blockages and reflections taking into account\cite{ref3}. These models were either limited to simple and inaccurate channel models or could not accurately reflect the characteristics of the channel in specific environments, which often result in large errors. Yet the models which use accurate data for channel modeling are impractical because of the high storage cost. In addition to modeling the physical environment model, some channel-related parameters are also needed.
\par
CKM is a site-specific database tagged with transceiver location pairs, which could give the real-time Channel State Information(CSI) acquisition and then be used to design UAV trajectory by turning uncertain information into certain ones. \textcolor{blue}{Precisely, uncertainty factors such as environment variations, location errors and interferences are considered and processed during the construction of CKM.} \textcolor{blue}{Through training with sufficient data, CKM provides accurate, efficient reflections of the channel’s characteristics, adapting seamlessly to dynamic conditions. Moreover, by continuously collecting and updating data within CKM, it effectively captures environmental changes in continuous 3D radio channels, making it especially valuable for 6G networks, which demand high adaptability, massive connectivity, and spectral efficiency.} CKM provides promising solutions to the practical issues arising from the significant expansion of channel dimensions and the associated training costs. 
Authors of \cite{ref5} presents a general framework for environment-aware communication leveraging CKM, and outlines several typical CKM-assisted communication scenarios. Similarly, the authors in \cite{ref6} optimized the accuracy of a radio map using a geostatistical tool named Kriging interpolation in cognitive radio networks. The authors in \cite{ref7} proposed an efficient RadioUNet for estimating the propagation pathloss learning from a physical simulation dataset.
Authors of \cite{ref8} established a 3D CKM to directly give the channel knowledge between the transmitter and the receiver to avoid the high delay and pilot interference caused by traditional pilot-based training. To place UAV in a UAV-aided relay scenario, the authors in \cite{ref9} constructed a full dimensional radio map to predict the channel gain between any transmitter location and any receiver location based on received signal strength measurements between low-altitude aerial nodes and ground nodes.
For the establishment of CKM, the most important things are the authenticity, reliability, and richness of the data. A well trained CKM can enhance UAV coverage with its accurate prediction and ensure robust communication links in various UAV-assisted applications. However, the data a UAV can collect during flight is often limited. The UAV's short flight time, limited storage and processing power, and the weight restrictions of sensors and other equipment reduce its overall performance in collecting data. As a result, relying only on the data collected by UAVs or from simulation software to train the CKM is time-inefficient. A key challenge is how to acquire additional data and effectively utilize the training data.
Besides, the data may be affected by errors resulting from satellite positioning inaccuracies,which frequently occurs due to factors such as shadowing, mobility, and time difference. When additional factors like obstacle reflection and propagation delay arise, these positioning errors are exacerbated, further affecting channel model estimation and diminish algorithm performance. To tackle this, in \cite{ref10}, a CKM is to trained to compensate the satellite positioning errors and provide more accurate channel gain information.
\par
AIGC refers to contents produced by AI in various forms, such as text, images, audio, and more. Notable advancements in this field include Generative Adversarial Networks(GANs)\cite{ref12}, diffusion models\cite{ref13}, and multimodal generation techniques\cite{ref14}. These technologies can be used for various fields, providing potential solutions to wireless communication system. To tackle the problem of limited data collection, we employed an AIGC technique, WGAN, \textcolor{blue}{to overcome the limitations of traditional data generation methods. By leveraging AI-generated content, the process can dynamically generate rich datasets that closely resemble real-world channel characteristics, thereby enhancing model-building processes and reducing the reliance on costly or difficult-to-obtain real-world data. By incorporating synthetic data through WGAN, we have significantly improved CKM’s computational efficiency and reduced data scarcity challenges, which is particularly beneficial in complex environments where traditional data collection is constrained.} \textcolor{blue}{Another use of AIGC is that, by integrating channel variations and environmental obstacles into CKM’s training, the model can be more accurate and better adapt to real-time changes in the communication environment. A CKM that effectively integrates such dynamic characteristics will be more resilient and capable of supporting the high-performance demands of UAV communication systems and enhances the accuracy and trustworthiness of predictions in real-world UAV scenarios.
}
\par
\par
\begin{figure*}[ht]
    \centering
    \includegraphics[width=0.8\linewidth]{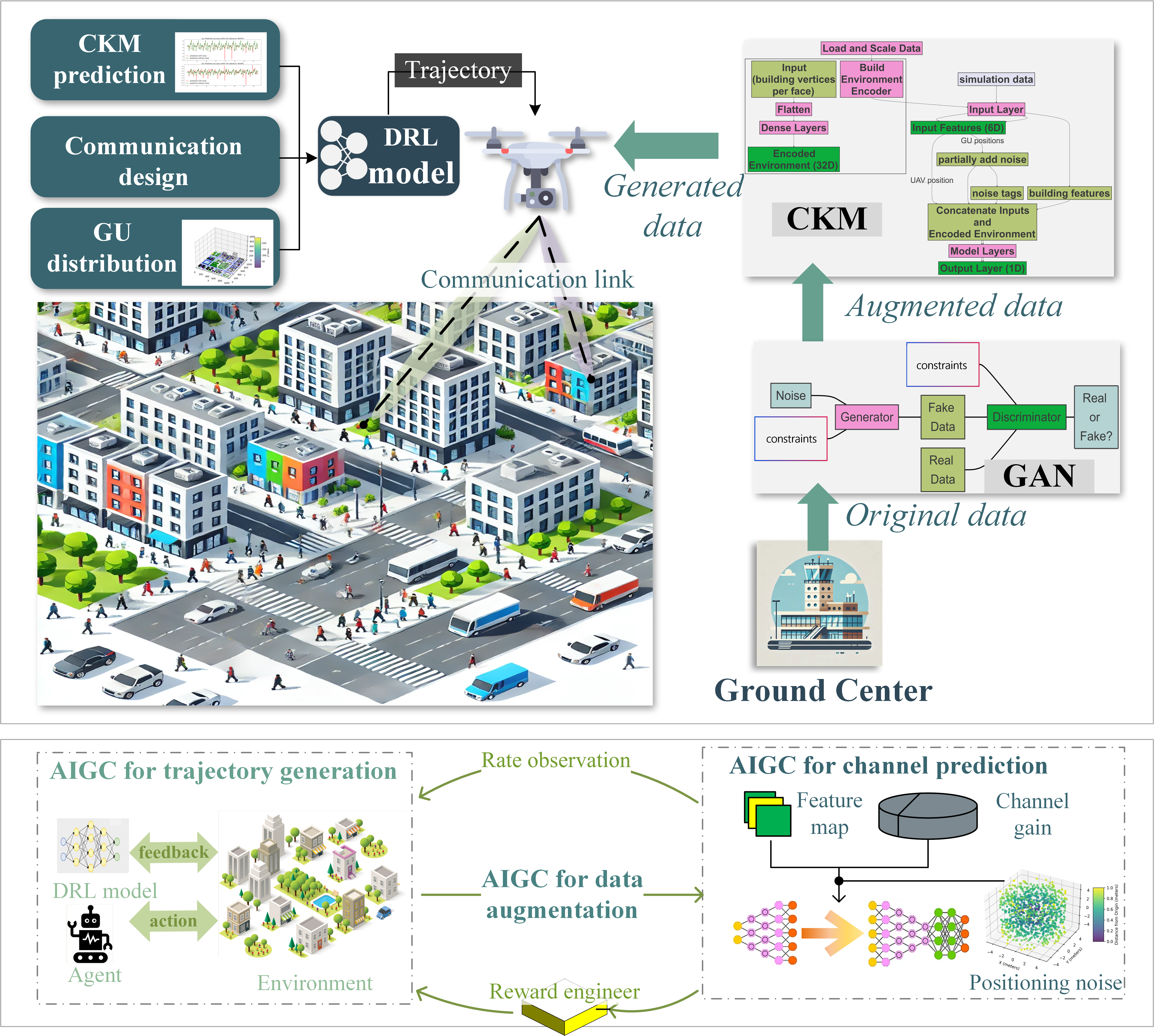}
    \caption{AIGC applications for communication design of the UAV-assisted system.}
    \label{fig1}
\end{figure*}

In this study, AIGC was employed to enhance data augmentation, channel prediction, and trajectory design, providing a viable solution to optimization problems that are difficult to address in industrial applications. By leveraging detailed environmental and channel information, simulations were conducted to gather diverse datasets for training a CKM that could adapt to dynamic urban environments. Then we integrated it into a UAV trajectory design framework, using Deep Reinforcement Learning (DRL) models to optimize communication links while taking into account various constraints such as transmitting power, channel conditions, and user distribution. The UAVs, serving as aerial communication nodes, generated a small time-cost trajectories that achieves communication demands for ground users. The CKM, coupled with WGAN data augmentation, helped overcome limitations in data collection, providing a more reliable solution for communication and resource management, as illustrated in Fig. \ref{fig1}.
\par
The contributions of this paper are summarized as follows.
\par
• A data augmentation scheme using WGAN was developed to expand the training dataset. By learning from the original dataset, the WGAN can produce realistic channel gain samples. This data augmentation process not only increases the volume of training data but also enhances the diversity of environmental scenarios, providing a more comprehensive dataset for training predictive models.
\par
• A CKM integrating channel knowledge into hidden layer was constructed to predict channel gain by leveraging the augmented dataset. The CKM was trained to map the relationship between user and UAV positions and the resulting signal attenuation. By incorporating both real and synthetically generated data, the CKM is capable of accurately predicting channel conditions across a wide range of environments. This model can support UAV communication by providing accurate, location-based signal predictions. 
\par
• A UAV trajectory design method utilizing the above AIGC was proposed to optimize communication paths in real-time. The CKM-generated channel gain data is integrated into a reinforcement learning framework, enabling the UAV to dynamically adjust its flight trajectory based on the predicted signal conditions. This method ensures that the UAV follows an optimal or near-optimal flight path, maximizing communication efficiency while minimizing signal loss.
\par
The remainder of this paper is organized as follows. Section II formulates the UAV trajectory and communication design problem. Section III displays the steps of AIGC application methods. Section IV presents the results analysis.

\section{Problem Formulation}
\subsection{The UAV-assisted Communication Framework}
Consider a wireless communication system where an UAV is deployed as an aerial base station to serve $N$ ground users (GUs) in a 3-dimensional space with a side of $X \times X \times H_{max}$. The UAV would fly in a set trajectory to optimize the communication links with the GUs and fly back to the GUs after communication. Each GU is assumed to have a fixed location on the ground, and the UAV aims to provide reliable and high-rate communication services to all GUs within its coverage area. The UAV's trajectory is designed to minimize the overall flying time, considering its hardware limitations, power constraints and communication demands.
\par
\begin{figure}[h]
    \centering
    \includegraphics[width=1\linewidth]{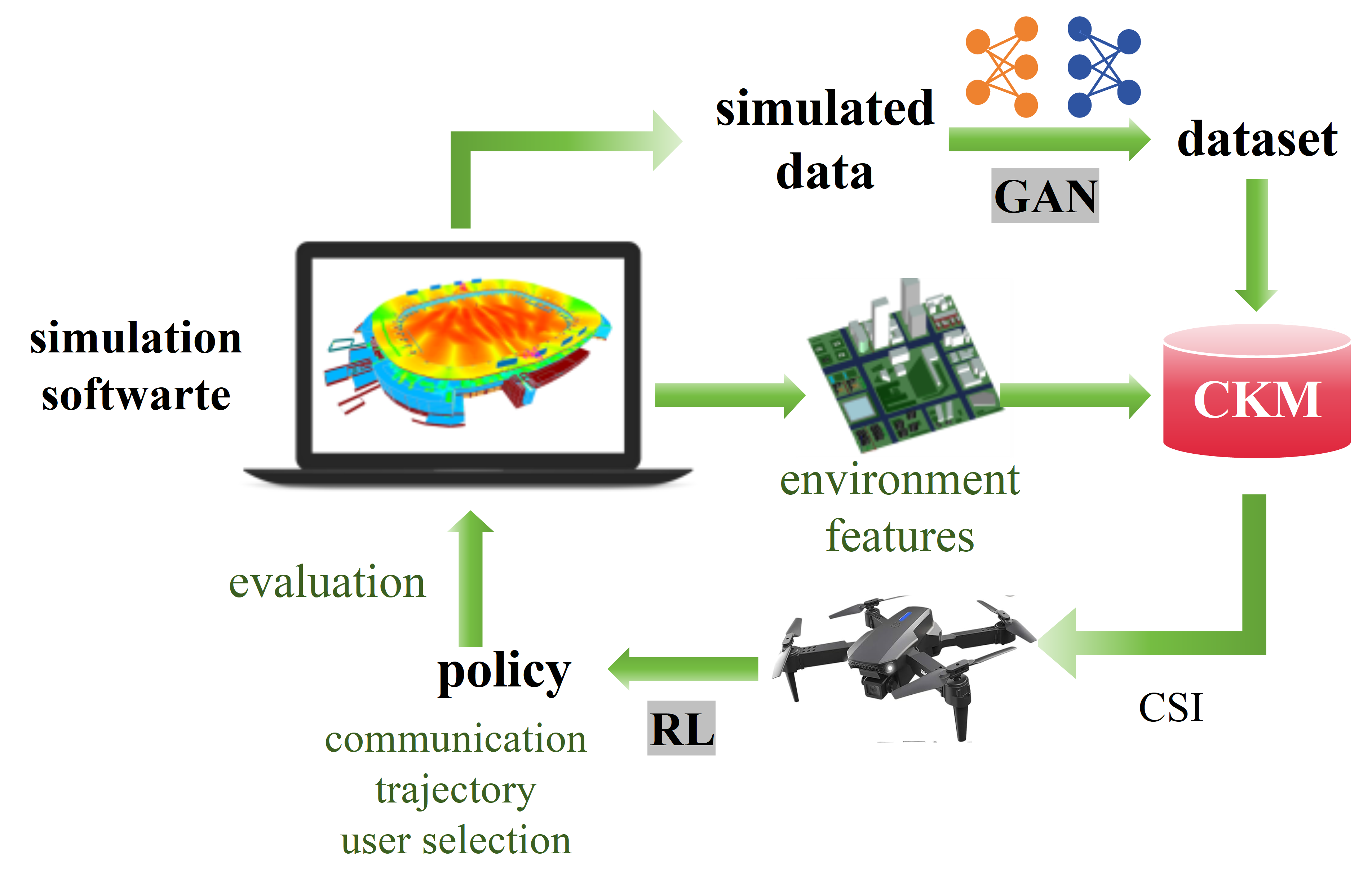}
    \caption{Validation of AIGC applications in UAV trajectory design.}
    \label{fig2}
\end{figure}

We assume that the UAV is flying above $H_{min}$, and the height of the buildings and the GUs is under it. As shown in Fig. \ref{fig2}, simulated data is generated and AIGC techniques are applied in data amplification and generation. Then the generated channel gains are used to calculate the throughput in RL environment to design a trajectory. 
\subsection{Channel Model}
Denote the location of the UAV by $\textbf{q}[t]=(x[t],y[t],z[t])\in \mathbb{R}^3$, and denote the location of the $n$-th GU by $\textbf{q}_n[t]=(x_n[t],y_n[t],z_n[t])\in \mathbb{R}^3$. Here, $0\leq t \leq T_{max}$. The original starting point of the UAV is $(0,0,H_{min})$.
\par
The wireless channel between the UAV and each GU is assumed to follow an A2G model, which is common in aerial communication scenarios. It could be modeled as a probabilistic summation of LoS and NLoS which could be both expressed as a basic free space pathloss plus an average shadowing:
\begin{equation}
    \hat{L}_X(d_n[t])=20log_{10}(\frac{4\pi f_cd_n[t]}{c})+\epsilon_X.
\end{equation}
$f_c$ is the carrier frequency, $c$ is the velocity of light and $d_n[t]$ is the distance of the $n$-th GU and the UAV which can be calculated by
\begin{equation}
    d_n[t]=||\textbf{q}[t]-\textbf{q}_n[t]||_2.
\end{equation}
Then according to the probability of LoS, the channel loss between the $n$-th GU and the UAV is
\begin{align}
    \hat{L}(\textbf{q}[t],\textbf{q}_n[t]) = &\ \frac{\epsilon_{LoS}-\epsilon_{NLoS}}{1+ae^{-b(tan^{-1}\frac{h_n[t]}{r_n[t]}-a)}} \\
    &\ +20log\frac{4\pi f_cd_n[t]}{c}+\epsilon_{NLoS}.
\end{align}
and
\begin{equation}
    r_n[t] = \sqrt{(x[t]-x_n[t])^2+(y[t]-y_n[t])^2},
\end{equation}
\begin{equation}
    h_n[t] = |z[t]-z_n[t]|,
\end{equation}

\par
To model the channel loss more accurately, a cite-specific CKM is trained to map the estimated channel loss into real loss, which could be expressed as $\mathcal{G}: [\hat{\mathcal{L}}(\mathcal{Q},\mathcal{Q}_n),\mathcal{F}_{env}] \xrightarrow{} \mathcal{L}(\mathcal{Q},\mathcal{Q}_n,\mathbf{E})$.
\par
Denote the transmitted power of the UAV by $P^T[t]$ and thus the received power of the $n$-th GU is
\begin{equation}
    P^r_n[t] = P^T[t] - L(\textbf{q}[t],\textbf{q}_n[t]).
\end{equation}
\par
The communication rate is thereby calculated by
\begin{equation}
    r_n[t] = B_n[t]log_{2}(1+\frac{P^r_n[t]}{\sigma^2}),
\end{equation}
$B_n[t]$ is the bandwidth allocated to the $n$-th GU at time $t$, and $\sigma^2$ represents the noise power.
\subsection{Formulated Problem}
The UAV's position in the inertial frame is determined by its velocity and orientation, which are functions of its roll ($\phi[t]$), pitch ($\theta[t]$), and yaw ($\psi[t]$) angles which describe its rotation angles around x, y, and z axis separately. The transformation from the body frame to the inertial frame is performed using the Direction Cosine Matrix (DCM), denoted as $\mathbf{C}_{BN}[t]$, which is expressed as:
\begin{equation}
\mathbf{C}_{BN}[t] = \mathbf{R}_z(\psi[t]) \mathbf{R}_y(\theta[t]) \mathbf{R}_x(\phi[t]),
\end{equation}
where $\mathbf{R}_x(\phi[t])$, $\mathbf{R}_y(\theta[t])$, and $\mathbf{R}_z(\psi[t])$ represent the rotation matrices about the $x$, $y$, and $z$ axes, respectively.
\par
The UAV's velocity vector in the inertial frame, $\mathbf{v}[t]$, then can be calculated as:
\begin{equation}
\mathbf{v}[t] = \mathbf{C}_{BN}[t] \cdot \mathbf{v}_B[t]
\end{equation}

where $\mathbf{v}_B[t]$ is the velocity vector in the body frame. The UAV's position $\mathbf{q}[t]$ is then obtained by integrating the velocity over time:

\begin{equation}
\mathbf{q}[t] = \mathbf{q}[t-1] + \mathbf{v}[t].
\end{equation}
\begin{equation}
    V[t] = V[t-1] + a[t],
\end{equation}
Here $V[t] = ||\textbf{v}[t]||$.
\par
Taking into account the speed constraints, the following conditions must be satisfied:
\begin{equation}
    -a^{max} \leq a[t] \leq a^{max},
\end{equation}
and 
\begin{equation}
    0 \leq V[t] \leq V_{max}.
\end{equation}

\par
The UAV is restricted to a maximum transmit power \(P_{max}\). However, the UAV does not necessarily operate at maximum power at all times, hence:
\begin{equation}
    0 \leq P^T[t] \leq P_{max}.
\end{equation}

\par
A threshold \(P_{min}\) is defined for GUs, ensuring that the \(i\)-th GU will only engage in communication if the received power \(P_i^R[t]\) exceeds \(P_{min}\); otherwise, it remains silent. Additionally, a binary association variable \(\alpha_i[t]\) is introduced to represent whether the \(i\)-th GU is in communication with the UAV, as defined by:
\begin{equation}  
    \alpha_i[t] = \left\{
        \begin{aligned}
        1, & \quad \text{if } P_i^R[t] \geq P_{min}, \\
        0, & \quad \text{otherwise}.
        \end{aligned}
        \right.
\end{equation} 

The following constraints are then:
\begin{equation}
    \alpha_i[t](P_i^R[t] - P_{min}) \geq 0,
\end{equation}
\begin{equation}
    \alpha_i[t] \in \{0,1\}.
\end{equation}   

\par
Given that the UAV functions as a base station, it must fulfill specific communication tasks. Each GU has a minimum required payload \(\eta_{min}\), which must be achieved by ensuring that the total throughput meets this requirement:
\begin{equation}
    \sum_{t=0}^{t=t_{end}} \alpha_i[t] R_i[t] \geq \eta_{min}.
\end{equation}

\par
The task is considered complete when either the maximum time \(T_{max}\) is reached or the required payload for each GU is delivered, at which point the UAV returns to its starting location. The goal is to minimize the total time from the start of the UAV's communication with the GUs until the completion of the task and the return to the starting point.

\par
The optimization problem involves the variables \(\alpha_i[t]\), \(\textbf{q}[t]\), and \(P_i^T[t]\). Among these, \(\textbf{q}[t]\) and \(P_i^T[t]\) are continuous variables, while \(\alpha_i[t]\) is binary. The problem can be mathematically expressed as:
\begin{small}
\begin{align} \label{13}
&\min_{\alpha_i[t],\textbf{q}[t],P_i^T[t]} \quad t_{end} \\
\text{s.t.}\quad
&t_{end} \leq t_{max} \tag{\ref{13}a}\\
&\textbf q[0] = \textbf q[t_{end}] = (0,0,H_{min}) \tag{\ref{13}b}\\
&-\frac{\pi}{2} \leq \theta[t] \leq \frac{\pi}{2}, \forall t = 0, 1, \dots, t_{end} \tag{\ref{13}c}\\
&-\pi \leq \phi[t] \leq \pi, \forall t = 0, 1, \dots, t_{end} \tag{\ref{13}d}\\
&-\pi \leq \psi[t] \leq \pi, \forall t = 0, 1, \dots, t_{end} \tag{\ref{13}e}\\
&0 \leq V[t] \leq V_{max}, \forall t = 0, 1, \dots, t_{end} \tag{\ref{13}f}\\
&-a^{max} \leq a[t] \leq a^{max}, \forall t = 0, 1, \dots, t_{end} \tag{\ref{13}g}\\
&0 \leq P^T[t] \leq P_{max}, \forall t = 0, 1, \dots, t_{end}, \forall i \in \mathcal{I} \tag{\ref{13}h}\\
&\alpha_i[t](P_i^R[t] - P_{min}) \geq 0, \forall t = 0, 1, \dots, t_{end}, \forall i \in \mathcal{I} \tag{\ref{13}i}\\
&\alpha_i[t] \in \{0,1\}, \forall t = 0, 1, \dots, t_{end}, \forall i \in \mathcal{I} \tag{\ref{13}j}\\
&\sum_{t=0}^{t_{end}} \alpha_i[t] R_i[t] \geq \eta_{min}, \forall i \in \mathcal{I} \tag{\ref{13}k}
\end{align}
\end{small}

\par
Upon task completion, the UAV must return to its initial position to recharge and be prepared for subsequent missions. In constraint (\ref{13}b), the initial time is set to 0, consistent with the typical simulation setup.

\par
This problem is characterized by high dimension (which increases with the number of GUs), time-variance, and non-convexity. Although Successive Convex Approximation(SCA) methods could be used to create convex approximations, particularly when dealing with the height variable coupled in the denominator, the complexity makes real-time solving challenging. Alternatively, DRL methods can circumvent the non-convexity by interacting with the Markov Decision Problem(MDP) environment to address the problem effectively.

\section{AIGC-Enhanced Environment-Aware UAV Trajectory Design}
This section explores several applications of AIGC in UAV-based industrial communication. First, due to the time-intensive nature of data collection, a GAN architecture is employed to enhance data augmentation processes. Subsequently, a CKM is developed to provide accurate predictions of channel gain. Building on these components, a DRL approach is applied to interact with the environment, enabling the generation of optimized UAV flight paths that adhere to communication constraints.
\subsection{AIGC for Simulation Data Augmentation}
Simulation data collection costs a lot of time, so a WGAN is trained in Algorithm 1 for augmentation to improve the diversity of the training data. The WGAN\cite{ref15} consists of a generator that produces synthetic data from random noise and a discriminator that estimates the Wasserstein distance between real and generated data. This distance helps to improve training stability and mitigates the mode collapse issue common in traditional GANs. The discriminator is trained to maximize this distance, while the generator aims to minimize it. Weight clipping ensures Lipschitz continuity, stabilizing the training process.

\begin{algorithm}[htbp]   
\small  
    \caption{WGAN Training}   
    \begin{algorithmic}  
        \REQUIRE{The positions of the UAV and GU and their corresponding fading.}  
        \REQUIRE{$latent\_dim$, $output\_dim$, $batch\_size$, $epochs$, $n_{critic}$, $clip\_value$, $lr$}  
        \STATE\textbf{Initialization:} Initialize generator $G$ and critic $D$ with random weights  
        \FOR{number of training iterations}  
            \FOR{$t=1, \dots, n_{\text{critic}}$}  
                \STATE Sample a batch of $m$ noise samples $\{z^{(i)}\}_{i=1}^m$ from prior $p_z(z)$  
                \STATE Sample a batch of $m$ real data samples $\{x^{(i)}\}_{i=1}^m$ from data distribution $p_{\text{data}}(x)$  
                \STATE Compute the critic loss: \\ 
                    $L_D = \frac{1}{m} \sum_{i=1}^m [D(x^{(i)}) - D(G(z^{(i)}))]
                     + \lambda \frac{1}{m} \sum_{i=1}^m \left(\|\nabla_{\hat{x}} D(\hat{x}^{(i)})\|_2 - 1\right)^2 $\\ 
                \STATE Update the critic by descending its stochastic gradient: \\ 
                    $\theta_D \leftarrow \theta_D - \alpha \nabla_{\theta_D} L_D $\\
            \ENDFOR  
            \STATE Sample a batch of $m$ noise samples $\{z^{(i)}\}_{i=1}^m$ from prior $p_z(z)$  
            \STATE Compute the generator loss:  \\
               $ L_G = -\frac{1}{m} \sum_{i=1}^m D(G(z^{(i)}))  $
            \STATE Update the generator by ascending its stochastic gradient: \\ 
                $\theta_G \leftarrow \theta_G + \alpha \nabla_{\theta_G} L_G $
        \ENDFOR  
        \ENSURE{Generated position pairs and the corresponding channel gains}
    \end{algorithmic}  
\end{algorithm}

\subsection{AIGC for Channel Gain Prediction}
CKM serves as a comprehensive database that stores pairs of transmitter-receiver (Tx-Rx) locations alongside their corresponding channel gains. To enhance the generalization capacity and applicability of the CKM, AIGC is integrated into its construction process. By leveraging AIGC, the CKM is capable of extrapolating and generating synthetic channel data that closely resembles real-world conditions, thereby addressing potential data sparsity and improving the robustness of the system across diverse environments and scenarios. This approach not only augments the CKM's predictive accuracy but also broadens its capacity to adapt to new or unseen Tx-Rx configurations.
\par
A knowledge-driven architecture serves as the foundational framework of the CKM, which integrates domain knowledge with data-driven techniques to enhance learning efficiency and accuracy. Within this architecture, the knowledge block is  characterized by the Line-of-Sight (LoS) probability channel to model the basic characteristics of the environment. Specifically, the knowledge block utilizes established propagation models to estimate the likelihood of a direct, unobstructed path between the transmitter and receiver. This LoS probability is calculated based on the distance, elevation angle, and physical obstructions in the environment, providing a prior estimation of signal attenuation.
\par
The knowledge block is integrated into the hidden layers of the CKM neural network to act as an auxiliary learning mechanism. During training, the outputs from this knowledge block are concatenated with the neural network’s internal features. By incorporating this physics-based knowledge, the network is better equipped to understand the interactions between the environment and signal propagation, which leads to more accurate predictions of channel characteristics.
The overall CKM architecture operates as a hybrid system that combines data-driven learning with physical modeling. The input features, such as the transmitter-receiver positions and environmental details, are fed into both the knowledge block and the neural network. The neural network captures the non-linear patterns in the data, while the knowledge block provides a structured, physics-guided representation of the environment’s impact on the signal. Together, these components enhance the predictive power of the CKM, ensuring that it not only fits the available data but also respects the underlying physical laws governing wireless communication.
\par
\begin{algorithm}
    \caption{Training of Knowledge Driven CKM}
    \begin{algorithmic}
        \REQUIRE{Dataset: User and UAV positions, Channel Gains, Environment Information}
        \REQUIRE{Parameters: A, a, b, B (for channel path loss model)}
        \REQUIRE{Model parameters: learning rate $\eta$, number of epochs $E$, batch size $B$}
        \STATE\textbf{Initialize} randomly split the dataset and then augment the training data
        \FOR{each epoch $e$ from 1 to $E$}
            \FOR{each mini-batch of size $B$}
            \STATE $\mathcal{Q}, \mathcal{L} \gets \text{randomly from mini-batch}$
            \STATE $\mathcal{F}_{env} \gets Enc(env)$
            \STATE $\textbf{x} \gets [\mathcal{Q},\mathcal{F}_{env}]$
            \STATE $\hat{\mathcal{L}} \gets {\mathcal{Q},A, a, b, B}$
            \STATE Apply knowledge block in hidden layer:
            \STATE $\Dot{\mathcal{H}} \gets \hat{\mathcal{L}}+res(\textbf{x})$
            \STATE Calculate the Mean Squared Error :
        \[
        \text{MSE} = \frac{1}{B} \sum_{i=1}^{B} (\Dot{L}_i - L_i)^2
        \]
        \STATE update the model's parameters:
        \STATE$\theta \overset{+}{\gets} -\eta \cdot \text{Adam}(\nabla_{\theta} \text{MSE})
        $
            \ENDFOR
        \ENDFOR
        \ENSURE{Trained neural network model for channel attenuation prediction}
    \end{algorithmic}
\end{algorithm}
\par
The Algorithm 2 explains how the network combines user and UAV positional data with environmental information to predict channel attenuation. The model takes two inputs: one for the user and UAV positions, and another for the environmental features. The environmental data is encoded into lower-dimensional features via a dedicated sub-network and then concatenated with the positional data for further processing. Multiple residual blocks are used to capture complex relationships between the inputs, with the final residual block incorporating a knowledge module based on a LoS (Line-of-Sight) probability path loss model. This knowledge enhances the model's ability to adjust predictions based on known path loss behaviors. After training, the model predicts channel attenuation on test data.

\subsection{MDP Formulation}
Utilizing the channel prediction from CKM, the resource allocation and trajectory design problem in Section II can be formulated as a MDP, where the goal is to minimize the total communication time \( t_{\text{end}} \), while satisfying the communication requirements of IoT devices and adhering to the physical limitations of the UAV. The MDP is characterized by the following components:
\par
\textbf{State Space \( \mathcal{S} \):} The state \( s[t] \in \mathcal{S} \) at each time step \( t \) is represented by a comprehensive vector that captures both UAV dynamics and communication status. The UAV state consists of its position \( \mathbf{q}[t] = (x[t], y[t], z[t]) \), velocity \( V[t] \), and orientation angles \( \theta[t] \), \( \phi[t] \) and $\psi[t]$, which describe the UAV's x-axis, y-axis and z-axis directions of movement. Additionally, each IoT device's state is represented by its position \( \mathbf{q}_{i}[t] = (x_{i}[t], y_{i}[t], z_{i}[t]) \) and communication status, which includes received power, path loss, and remaining data to be transmitted. The complete state vector also includes whether a communication link has been established with the IoT device, as well as environmental conditions such as noise and the UAV's path loss.

\textbf{Action Space \( \mathcal{A} \):} At each time step, the UAV takes an action \( a[t] \in \mathcal{A} \), which determines its movement and transmission power. The action consists of four components: the UAV's velocity $V[t]$, three dimensional directions \( \theta[t] \), \( \phi[t] \) and $\psi[t]$ and transmitting power $P_T[t]$. These control variables are constrained by the UAV's physical limits, such as its maximum velocity \( V_{\text{max}} \) and maximum transmission power \( P_{\text{max}} \). 

The action space is bounded such that the UAV can only select velocities and angles within the feasible limits, as described by the problem constraints.

\textbf{Transition Model \( P(s_{t+1} | s_t, a_t) \):} The state transition dynamics are governed by the UAV's movement and communication decisions. At each time step, given the current state \( s[t] \) and action \( a[t] \), the UAV's position is updated based on its velocity and direction. The IoT device's communication status is updated depending on the received power and path loss between the UAV and the device. The transition model also incorporates environmental noise and uncertainties in the channel model, which affect the communication success.

\textbf{Reward Function \( r(s_t, a_t) \):} The reward function encourages efficient communication while penalizing excessive movement or poor decisions. The reward at each time step is composed of several terms:
- \( r_1 \): Penalty for inefficient UAV movement (e.g., moving too far away from IoT devices or exceeding the allowed velocity).
- \( r_2 \): Reward for successfully transmitting data to an IoT device, which is proportional to the amount of data transmitted.
- \( r_3 \): Bonus for completing the communication task early.
- \( r_4 \): Penalty for running out of time without completing the communication task.
\par
\textbf{Objective:} The objective of the MDP is to find an optimal policy \( \pi^*(s_t) \) that minimizes the total communication time \( t_{\text{end}} \), while satisfying the constraints of trajectory limits, velocity bounds, acceleration, and minimum required data transmission for each IoT device. The solution to this MDP determines the optimal control strategy for the UAV's movement and communication decisions, ensuring minimal communication time while maintaining system feasibility.
\begin{algorithm}[htbp]
 \small
\begin{algorithmic}
\caption{Training of UAV Trajectory and Communication Optimization}
\REQUIRE{Environment: UAVs, User Positions, Channel Gains, Environment Information}
\REQUIRE{Parameters: learning rate $\eta$, number of episodes $E$, max time $T_{max}$}
\REQUIRE{PPO Hyper-parameters: clipping range $\epsilon$, discount factor $\gamma$, policy and value network architecture}

\textbf{Initialize} environment, neural network policy $\pi_{\theta}(a|s)$, and value function $V_{\phi}(s)$\;\\
\textbf{Initialize} Adam optimizer for both $\pi_{\theta}(a|s)$ and $V_{\phi}(s)$\;\\

\FOR{episode $e$ from 1 to $E$}
    \STATE \textbf{Reset} environment to obtain initial state $s_0$\;
    \STATE \textbf{Initialize} empty lists to store trajectories: states, actions, rewards, and log probabilities\;

    \FOR{time $t$ from 1 to $T_{\text{max}}$}
        \STATE \textbf{Sample} action $a_t \sim \pi_{\theta}(a|s_t)$\;
        
        \STATE \textbf{Select user}: Use action $a_t$ to select several users from the environment:
            \[
            \mathcal{L} \gets \mathcal{G}
            \]
        \STATE \textbf{Calculate power}: Calculate the communication power for the selected user;
        \STATE \textbf{Update environment}: Apply the action to update the state $s_{t+1}$, reward $r_t$, and done flag\;
        \IF{done}
            \STATE \textbf{Break} loop
        \ENDIF
    \ENDFOR
    
    \FOR{mini-batch}
        \STATE \textbf{Calculate} advantage estimates $A_t$ using GAE:\\
        \[
        A_t = \sum_{t'=t}^{T} \gamma^{t'-t} \left( r_{t'} + \gamma V_{\phi}(s_{t'+1}) - V_{\phi}(s_{t'}) \right)
        \]\\
        \STATE \textbf{Optimize} policy $\pi_{\theta}(a|s)$ and value function $V_{\phi}(s)$ using PPO updates\;\\
        \STATE \textbf{Clip} the ratio $r(\theta) = \frac{\pi_{\theta}(a|s)}{\pi_{\theta_{\text{old}}}(a|s)}$ to avoid large updates:\\
        $\mathcal{L}^{\text{CLIP}}(\theta) = \min \left( r(\theta) A_t, \text{clip}(r(\theta), 1 \pm \epsilon) A_t \right)$

        \STATE \textbf{Update} model parameters:\\
        \[
        \theta\overset{+}{\gets} - \eta \nabla_{\theta} \mathcal{L}^{\text{CLIP}}
        \]\\
    
    \ENDFOR

\ENDFOR
\ENSURE{Trained policy network $\pi_{\theta}(a|s)$ and value function $V_{\phi}(s)$ for optimizing UAV trajectories and communication efficiency} 
\end{algorithmic}
\end{algorithm}
\subsection{AIGC for UAV Trajectory and Communication Design}
An AIGC technique-Reinforcement learning plays a significant role in optimizing UAV trajectories to enhance decision-making and efficiency. PPO is such a typical policy gradient method designed to improve upon traditional reinforcement learning algorithms by striking a balance between exploration and exploitation. Specifically, the PPO algorithm iteratively updates the UAV's policy to maximize a cumulative reward function, which, in this case, is designed to enhance communication performance while minimizing flight costs and energy consumption.
\par
The process begins by defining a neural network policy with several hidden layers that map the UAV’s observation space—comprising user positions, obstacles, and communication requirements—into a sequence of actions that control the UAV’s movement. The policy network is optimized using a set of rewards and penalties related to the communication link quality, distance to GUs, and the avoidance of physical obstacles. Through repeated interaction with the environment, the algorithm improves the UAV’s trajectory by adjusting its actions to maximize the expected cumulative reward, while constraining the updates to avoid drastic changes that might destabilize learning.
\par
AIGC augments the training environment by generating high-quality synthetic data that captures environmental variability. This enriched dataset enables the PPO algorithm to learn more robust UAV control policies, especially in scenarios where actual channel data is sparse or difficult to collect.
\par
The use of PPO, combined with the data augmentation provided by AIGC, allows for an optimized balance between communication quality and resource efficiency, making it a powerful solution for UAV trajectory design in dynamic environments.
As shown in Algorithm 3, the CKM is utilized in the step phase to obtain the prediction of the channel gains in each episode and then the prediction is used to calculate the throughput of the system.

\section{Simulation Results}
In this section, simulation experiments are conducted to sequentially evaluate the effects of AIGC-enabled data augmentation, channel prediction, and trajectory design.
\par
In data augmentation phase, a WGAN is trained using the dataset composed of real channel attenuation data, which includes the UAV and GU locations with corresponding attenuation values. We use this dataset as the foundation for generating synthetic channel data. The generator network is initialized with high dimensional random noise vectors as inputs and is designed to output data samples that simulate the distribution of the real channel data. The discriminator network evaluates the generated samples by estimating the Wasserstein distance between the real and synthetic data distributions.
The training process alternates between updating the generator and discriminator, with the discriminator trained 3 times for every generator update to ensure convergence. A batch size of 256 is used, and the Adam optimizer is employed for both networks with a learning rate of 0.0001. Weight clipping is set to a threshold of 0.01 to maintain Lipschitz continuity in the discriminator.
\par
In the channel prediction phase, a custom neural network was employed to predict channel attenuation by leveraging a combination of raw input features, environment-specific data. The input dataset comprised both original and augmented features, including UAV and user coordinates normalized to the scene's size. And channel gains were normalized using the min-max method. The augmented dataset combined original input features with generated data.
A custom architecture was developed to integrate environmental features through a dedicated encoding module, which extracted relevant information from a 3D model of the environment, including building positions, heights, and structural details. The environment encoder consisted of dense layers followed by batch normalization to speed up convergence. The main network architecture employed residual blocks, with Dense layers of sizes 512, 256, 128, and 64, using ReLU activation and a regularization factor.
The model was trained using the Adam optimizer with an adaptive learning rate. Early stopping was implemented, ceasing training after 10 epochs without improvement in validation loss. The model was trained for up to 500 epochs, with the dataset split 70:30 for training and validation.
\par
To better utilize channel characteristics, a generalized A2G channel knowledge is integrated into two networks, forming a knowledge-featured CKM(KF-CKM) and a knowledge-driven CKM(KD-CKM). In KF-CKM model, an additional path loss feature is computed based on the distance and angle between the UAV and user using the eq.(3). In the KD-CKM model, a knowledge module based on the loss model is embedded directly into the hidden layers. The knowledge module modifies the hidden layers by providing real-time channel information. This ensures that the model learns not just from the raw data but also from a theoretically grounded understanding of channel attenuation.
\par
The simulation is set in a 1000m $\times$ 1000m $\times$ 750m area with UAV heights ranging from 250 m to 750 m, and the users are randomly positioned at a height of 250 m. The scenario incorporates 15 GUs as users with communication demands. The UAV begins at the coordinates (0, 0, 250), and its maximum velocity is set to 50 m/s, with a maximum acceleration of 20 m/s². The max transmission power is 33 dBm, adjusting dynamically by the algorithm to optimize communication efficiency based on the channel conditions, and the communication threshold is -70dBm.
The reward function guiding the UAV's behavior is designed to balance efficiency and task completion. It applies a penalty for excessive movement, while offering a reward for successfully communicating with a new user. If the UAV completes communication tasks ahead of the allocated time, it receives an additional reward for early completion. There is also a penalty if the UAV’s elevation angle drops below a critical threshold. The total reward is a weighted summation of them and it will ensure the UAV prioritizes efficient communication while maintaining optimal trajectories.
\par
To detect the quality of augmented data, a pairplot visual tool is applied to analyze relationships between different features in the dataset. Diagonal plots show the distribution of individual variables, providing insight into the spread and range of each feature. Off-diagonal scatter plots highlight the relationships between each pair of variables, revealing patterns, correlations, and possible outliers.
\begin{figure}[h]
    \centering
    \includegraphics[width=1.2\linewidth]{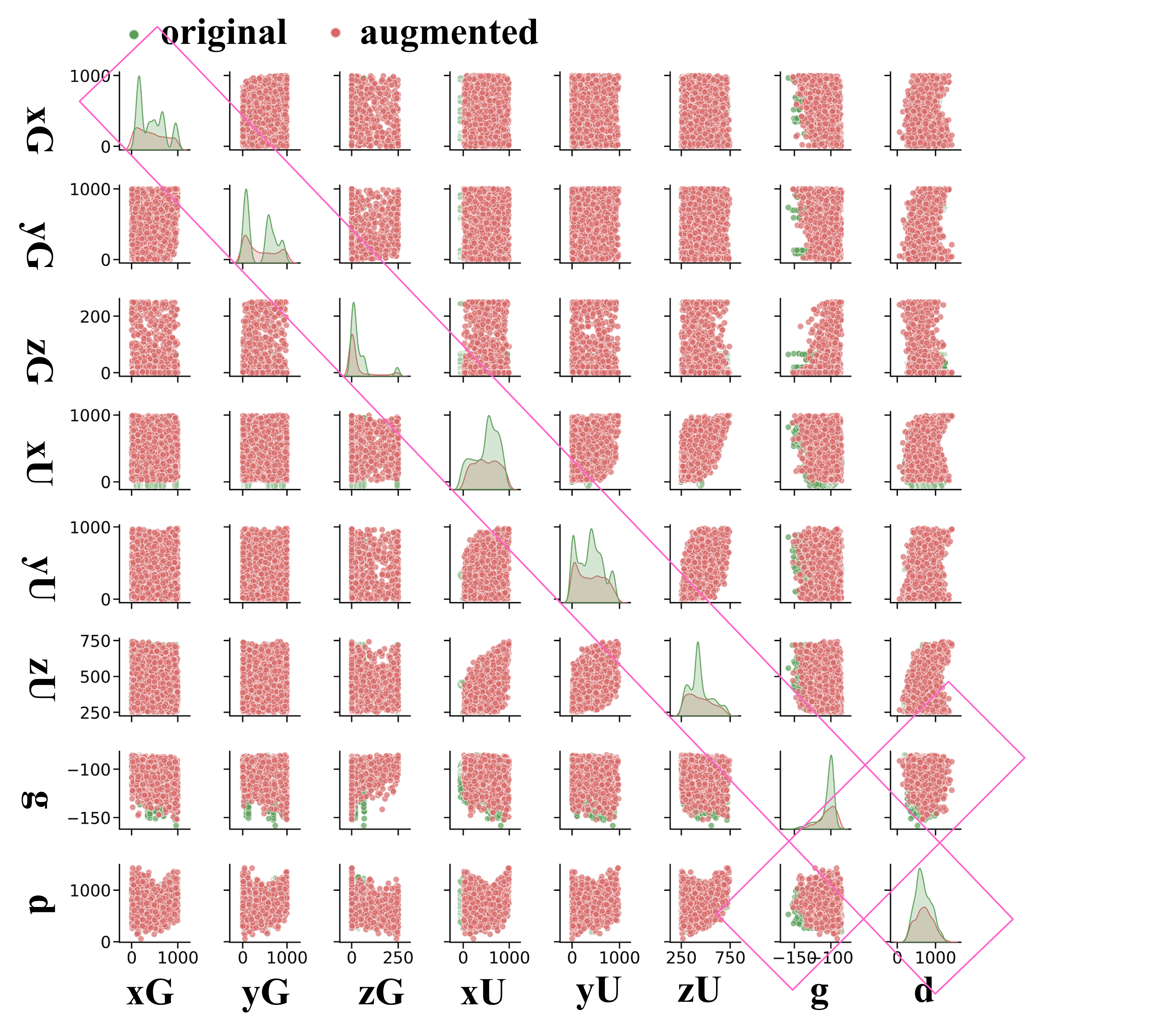}
    \caption{Distribution visualization of original and augmented data, with 'xG, yG, zG' for GUs' positions, 'xU, yU, zU' for GUs' positions, 'd' for distance between them and 'g' for the channel gains. The pink boxed section contains distribution plots for each variable as well as the relationship plots between distance and gain.}
    \label{fig3}
\end{figure}
It is shown from Fig. \ref{fig3} that the generated data maintains coordinates within the specified range, with a distribution closely aligned to that of the original data, indicating that the augmentation process effectively preserves key spatial characteristics. A critical aspect of the evaluation is the channel gain, which remains consistent with the original dataset, demonstrating that the generated data accurately captures channel attenuation patterns. The relationship between channel gain and the distance between the UAV and the user exhibits a negative correlation, as expected. Notably, this correlation is non-linear, since channel gain is influenced by both distance and elevation angle, reflecting the complex interactions defined in the channel model. The generated data successfully captures the non-linear dependencies between channel gain, distance, and elevation angle, illustrating the quality of AIGC.
\par

\par
\begin{figure}[t]
    \centering
    \includegraphics[width=1\linewidth]{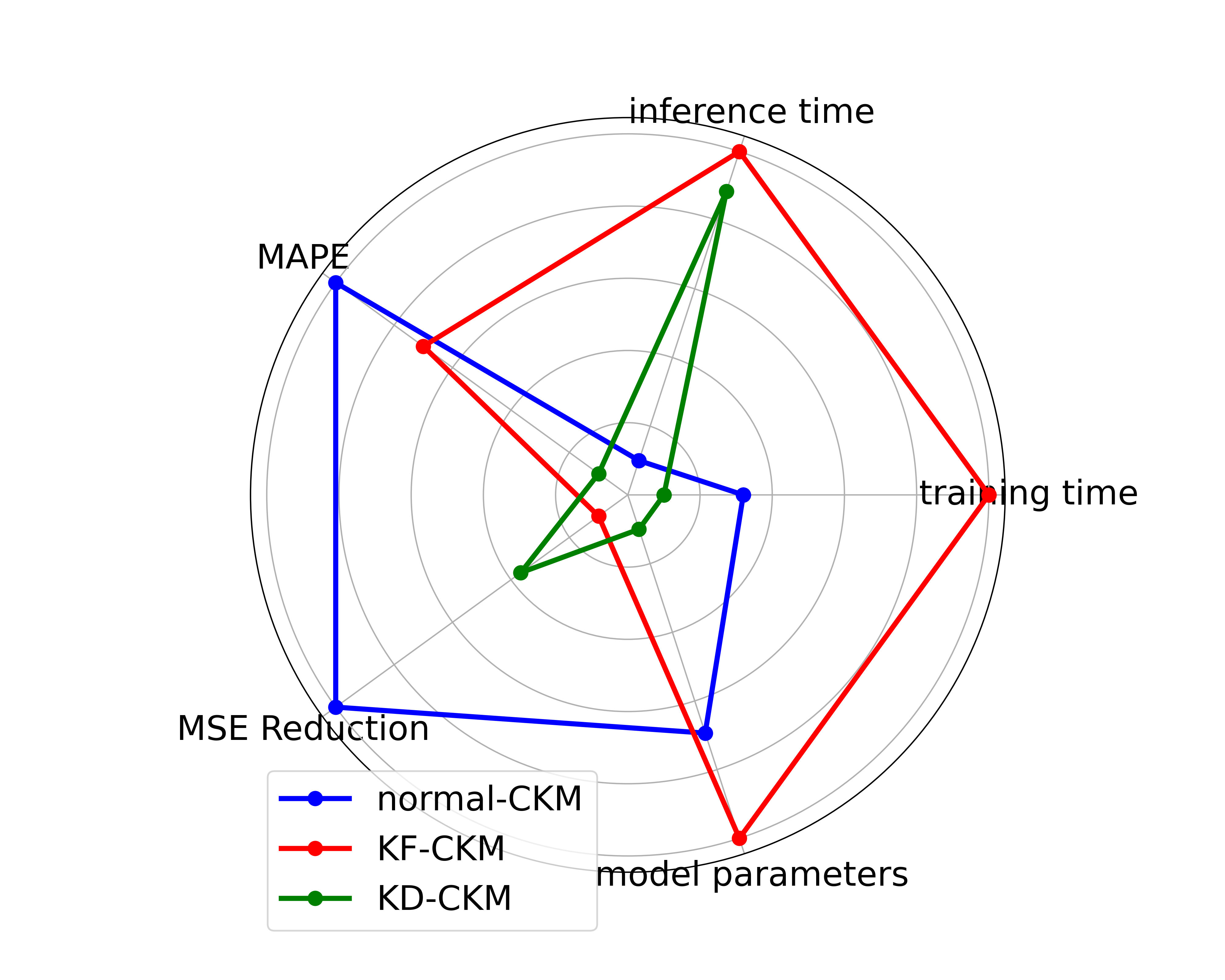}
    \caption{Radar plot of five parameters to assess the networks: training time, inference time, MAPE, MSE Reduction, model parameters\textcolor{blue}{(lower values indicate better performance for all metrics except MSE reduction)}.}
    \label{fig4}
\end{figure}
\par
\begin{figure*}[t]
    \centering
    \includegraphics[width=0.6\linewidth]{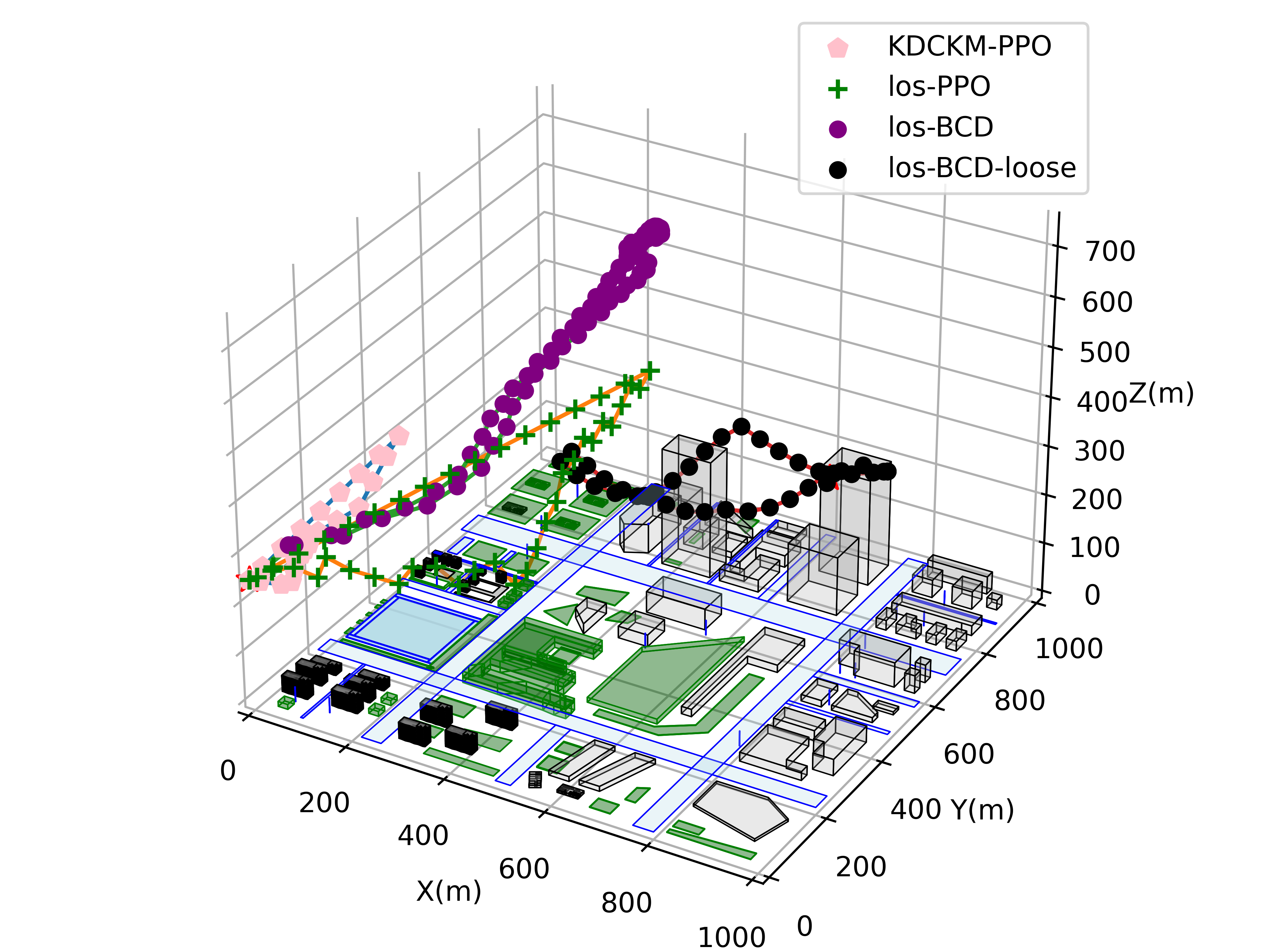}
    \caption{Generated Trajectories Comparison of Four methods. Los-PPO: PPO algorithm with LoS model; KDCKM-PPO: PPO algorithm with KD-CKM; los-BCD: BCD algorithm with LoS model; los-BCD-loose: los-BCD with start point not fixed.}
    \label{fig6}
\end{figure*}
Furthermore, AIGC technology and the data amplified in the previous step are used to train CKM for channel prediction and the performance of three networks are compared. In Fig. \ref{fig4} several parameters are chosen to be evaluated: \textcolor{blue}{training time, inference time, MAPE, and model parameters are minimization metrics, where lower values indicate improved efficiency or accuracy. MSE reduction measures the improvement brought by the enhanced model, with larger reductions indicating better predictive accuracy.}. In order to unify these parameters in one radar chart, they are normalized using max-min scaling to the range of 0.1 to 1. 
The longest training time is for normal KF-CKM since using simple signal gain as the primary feature increased training time, as the model had to learn from scratch. However, embedding domain knowledge, such as path loss in the hidden layers, significantly improved convergence speed by helping the network better capture feature relationships. The basic CKM model had the shortest inference time due to its simpler architecture, which required fewer computations. MAPE performance was good for both KF-CKM and KD-CKM, likely because incorporating domain knowledge helped reduce large prediction errors. MSE reduction(30.23\%) was most notable in the normal CKM model, which initially had the lowest accuracy, allowing data augmentation to have a more pronounced effect. The number of model parameters reflected the complexity of each network, with KD-CKM and KF-CKM having more parameters due to their advanced architectures.
\par
The results clearly show that AIGC is valuable in making good prediction, especially that the KD-CKM benefits the most from both the availability of original data and data augmentation, consistently achieving lower MSE across all conditions. The augmentation technique leads to noticeable improvements across all algorithms, highlighting its importance in improving CKM-based methods.
\par
Utilizing the KD-CKM for dynamic GU selection and channel gain prediction, DRL is applied to solve this optimizing problem. Three approaches are compared: DRL with CKM-based channel prediction, RL using a LoS probability model, and a Block Coordinate Descent(BCD) algorithm with the LoS model. The minimal flight time is shown in Table I and the trajectories are displayed in Fig. \ref{fig6}.
\begin{table}[h!]
  \begin{center}
    \caption{Flight time of four algorithms}
    \begin{tabular}{c|c|c} 
      \textbf{Algorithm} & \textbf{Flight Time(s)} & \textbf{Throughput(bps)}\\
      \hline
      los-BCD & 81 & 7.16 \\
      \hline
      los-BCD-loose & 50 & 11.62 \\
      \hline
      los-PPO & 45 & 14.97 \\
      \hline
      KDCKM-PPO & 37 & 16.18 \\
    \end{tabular}
  \end{center}
\end{table}
\par
From Fig. \ref{fig6} it can be observed that the trajectory derived from the BCD method tends to exhibit more vertical fluctuations as it seeks to achieve a higher LoS probability, balancing between maximizing coverage and minimizing path loss. In contrast, the AI-based algorithm, with its ability to perform global search, produces a more direct and efficient path. Therefore, the system throughput is higher. Trajectories computed using the LoS probability formula for channel attenuation tend to remain higher in altitude overall, aiming for better coverage. Moreover, another key advantage of AI is that, BCD is prone to getting stuck in local optima, and in some cases, may even fail to converge to a solution. The problem constraints we posed sometimes restrict initial conditions, leading to failed solutions. Only by relaxing these constraints can the BCD method approximate the AI-generated trajectories, but the resulting paths often no longer meet our expectations.
Besides, AIGC offers superior generalization: while traditional methods must re-optimize for new scenarios, AI dynamically adapts to changing conditions, providing a more robust and efficient approach for UAV communication optimization.
\section{Conclusion}
This paper presents several robust applications of AIGC in UAV communication. It applies the AI generated data that have similar distribution with the simulated data to improve the performance of subsequent steps. And then a novel approach for constructing a CKM that significantly enhances prediction accuracy by leveraging channel knowledge is introduced, which is then applied to optimize UAV communication resources with AI. Experimental results demonstrate superior performance of AIGC in both channel prediction and trajectory generation compared to traditional algorithms. This innovative approach not only improves the efficiency of UAV trajectory design but also offers valuable insights for future industrial applications. The power of AIGC is evident in the ability to process complex data and generate optimal solutions of neural networks, outperforming conventional techniques. Future research could explore extending the adaptability and real-time performance of AI models to enhance their robustness across dynamic and large-scale environments.

\newpage

\section*{Author Biography}
\vspace{}
\begin{IEEEbiography}
[{\includegraphics[width=1in,height=1.25in,clip,keepaspectratio]{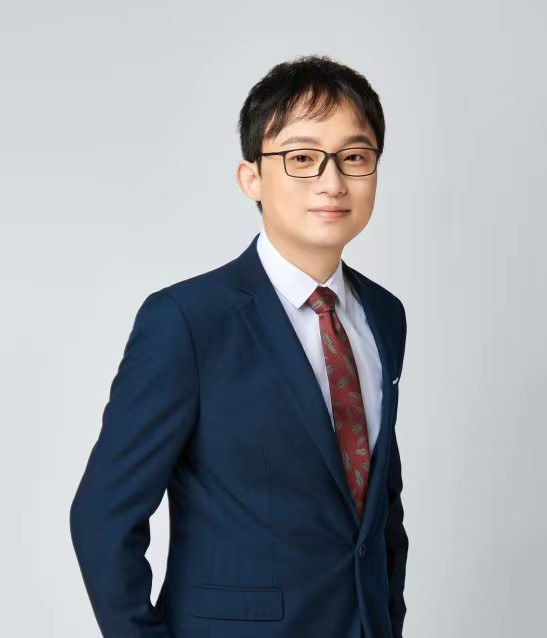}}]{Chiya Zhang}
received the Ph.D. degree in Telecommunication Engineering from the University of New
South Wales, Sydney, Australia, in 2019. He is currently an Associate Professor at Harbin Institute of
Technology, Shenzhen, China. His current research
interest is AI applications in telecommunication engineering. He received the Exemplary Reviewer Certificates of the IEEE Wireless Communications Letters in
2018 and IEEE ComSoc Asia-Pacific Outstanding Paper Award in 2020. He is serving as an Editor for the IEEE Internet of Things
Journal.
\end{IEEEbiography}
\vspace{}
\begin{IEEEbiography}
[{\includegraphics[width=1in,height=1.25in,clip,keepaspectratio]{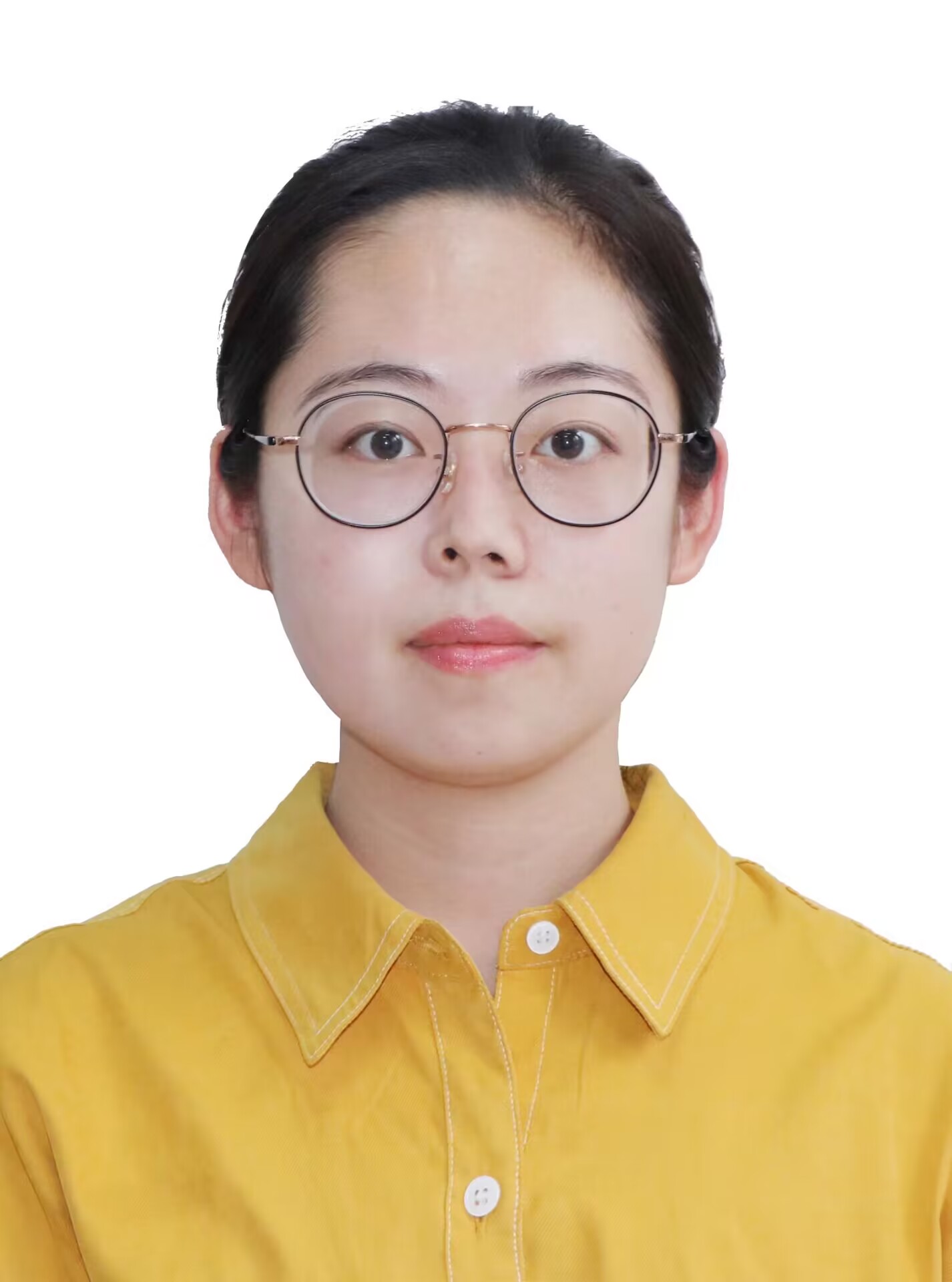}}]{Ting Wang}
received the B.S. degree in Communication Engineering from Harbin Institute of Technology, Shenzhen, China, in 2022. She is currently pursuing her Ph.D. degree in the School of Electronic and Information Engineering at the same institution. 
Her current research interest is the resource optimization in 6G, Channel Knowledge Map, and space-air-ground integrated network.
\end{IEEEbiography}
\vspace{}
\begin{IEEEbiography}
[{\includegraphics[width=1in,height=1.25in,clip,keepaspectratio]{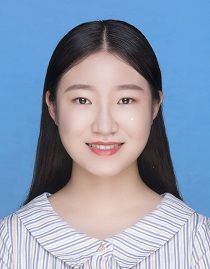}}]{Rubing Han}
received the B.S degree in Telecommunication Engineering form Harbin Institute of Technology, Shenzhen, China, in 2023. She is presently pursuing her Master's degree in the School of Electronic and Information Engineering at the same institution. She currently focuses on the design of UAV trajectories and the application of deep reinforcement learning.
\end{IEEEbiography}
\vspace{}
\begin{IEEEbiography}
[{\includegraphics[width=1in,height=1.25in,clip,keepaspectratio]{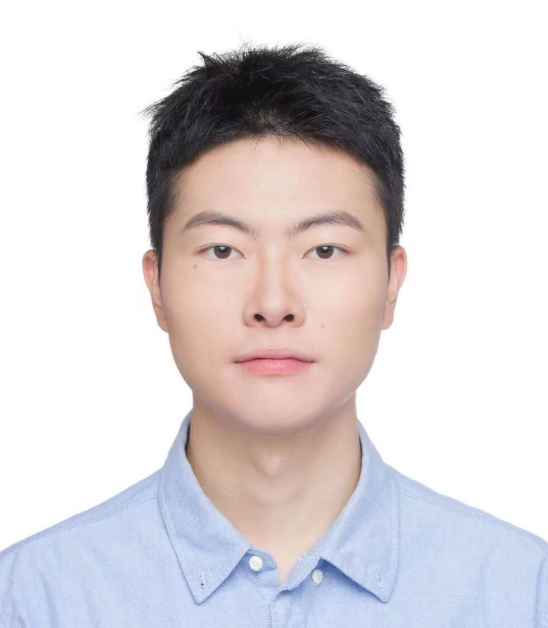}}]{Yuanxiang Gong}
received the B.S. degree in
Telecommunication Engineering from Harbin Institute
of Technology, Shenzhen, China, in 2023. He is currently a Master’s student at the School of Electronic
and Information Engineering, Harbin Institute of Technology, Shenzhen, China. His research interests include image processing, cybersecurity, and deep learning.
\end{IEEEbiography}

\end{document}